\documentclass{article}
\usepackage{ijcai23}

\usepackage{times}
\usepackage{soul}
\usepackage{url}
\usepackage[hidelinks]{hyperref}
\usepackage[utf8]{inputenc}
\usepackage[small]{caption}
\usepackage{graphicx}
\usepackage{amsmath}
\usepackage{amsthm}
\usepackage{booktabs}
\usepackage{algorithm}
\usepackage{algorithmic}
\usepackage[switch]{lineno}
\usepackage{amssymb}
\usepackage{booktabs}
\usepackage{multirow}
\usepackage{makecell}
\usepackage{bm}
\usepackage{appendix}
\usepackage{color}
\usepackage{xcolor,soul}

\title{From Association to Generation: Text-only Captioning by Unsupervised Cross-modal Mapping}

\author{
Junyang Wang$^1$\thanks{Work done during internship at DAMO Academy, Alibaba Group.}
\quad
Ming Yan$^2$\quad
Yi Zhang$^1$\quad
Jitao Sang$^1$\thanks{Corresponding author}
\affiliations
$^1$School of Computer and Information Technology, Beijing Jiaotong University\\
$^2$DAMO Academy, Alibaba Group\\
\emails
\{junyangwang, yi.zhang, jtsang\}@bjtu.edu.cn,
ym119608@alibaba-inc.com
}

\begin{document}

\maketitle

\begin{abstract}
    With the development of Vision-Language Pre-training Models (VLPMs) represented by CLIP and ALIGN, significant breakthroughs have been achieved for association-based visual tasks such as image classification and image-text retrieval by the zero-shot capability of CLIP without fine-tuning. However, CLIP is hard to apply to generation-based tasks. This is due to the lack of decoder architecture and pre-training tasks for generation. Although previous works have created generation capacity for CLIP through additional language models, a modality gap between the CLIP representations of different modalities and the inability of CLIP to model the offset of this gap, which fails the concept to transfer across modalities. To solve the problem, we try to map images/videos to the language modality and generate captions from the language modality. In this paper, we propose the \textbf{K}-\textbf{n}earest-ne\textbf{igh}bor Cross-modali\textbf{t}y Mapping (Knight), a zero-shot method from association to generation. With text-only unsupervised training, Knight achieves state-of-the-art performance in zero-shot methods for image captioning and video captioning. Our code is available at \textcolor{blue}{\url{https://github.com/junyangwang0410/Knight}}. 
\end{abstract}
\section{Introduction}
In the development of multi-modal learning, two recursive levels arise: (1) multi-modal association; (2) cross-modal generation. The former relies on multi-modal inputs and calculates the association scores for given multi-modal inputs through association expressions. Typical tasks include image classification, image-text retrieval, object detection, etc. The latter is to convert the input from one modality to other modalities, which requires a cross-modal transformation relationship to ensure that the same concept within different modalities can be represented accurately. Typical tasks include image-to-text and text-to-image generation.

Currently, Vision-Language Pre-training Models (VLPMs) represented by CLIP~\cite{radford2021learning} and ALIGN~\cite{jia2021scaling} have been successful at the first level by advantage of the multi-modal association-based pre-training task: contrastive learning. With 400 million massive (image, text) training data, CLIP successfully models the association between vision modality and language modality. Benefiting from the diverse web data, CLIP has an extensive perception of open-world knowledge. Without fine-tuning, CLIP achieves an accuracy of 76.2$\%$ on ImageNet in the zero-shot setting and rivals or even surpasses the fine-tuning model on datasets with multiple domains~\cite{radford2021learning}. Many multi-modal association-based works have benefited from the powerful zero-shot capability of CLIP: not only avoids the high collection cost of supervised data but also simplifies the deployment process.

\begin{figure}[t]
\centering
\includegraphics[width=0.45 \textwidth]{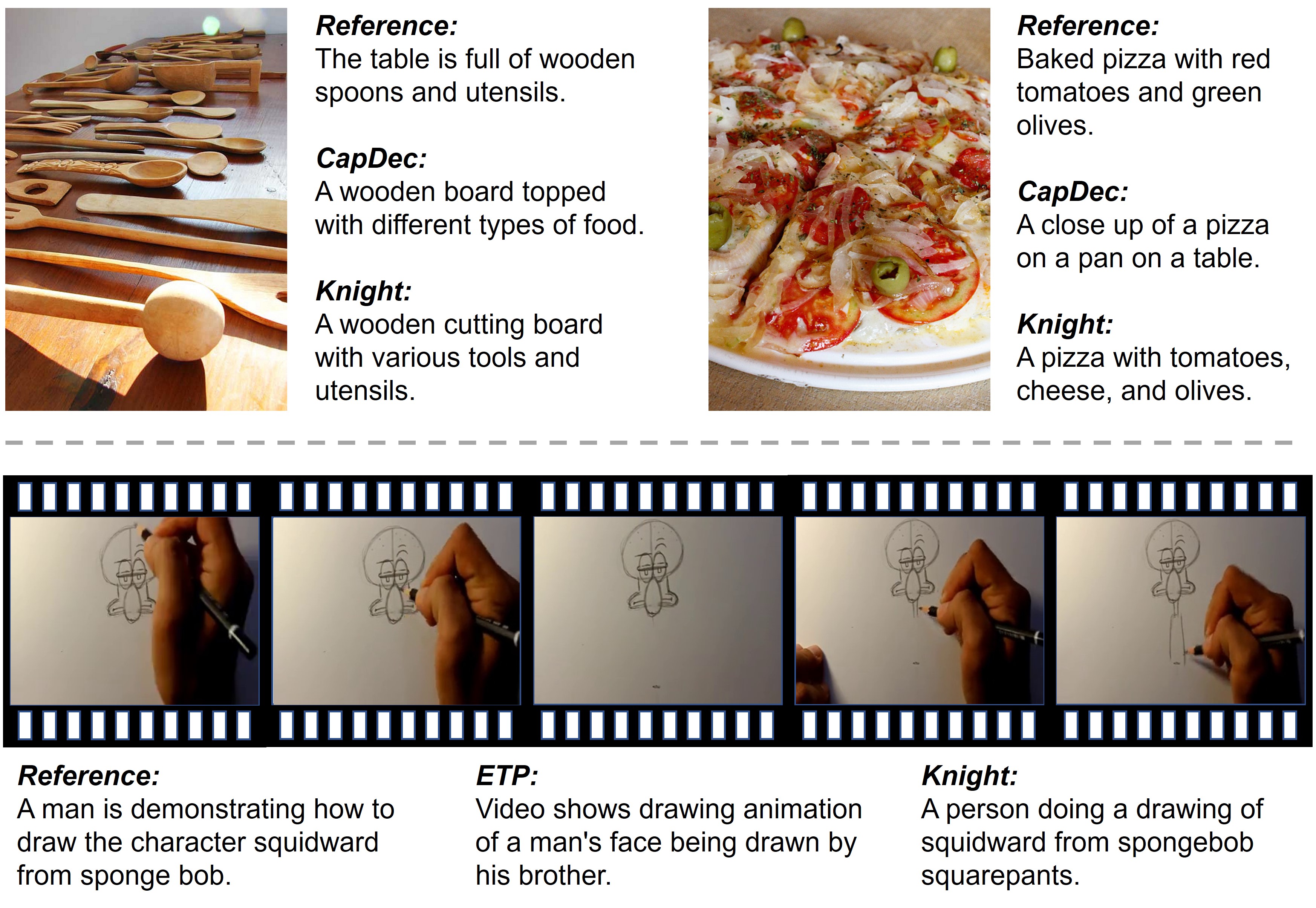}
\caption{\label{fig:figure0} The examples of \emph{Knight} compared with current state-of-the-art text-only captioning methods.}
\vspace{-4mm}
\end{figure}

\begin{figure*}[t]
\centering
\includegraphics[width=0.9 \textwidth]{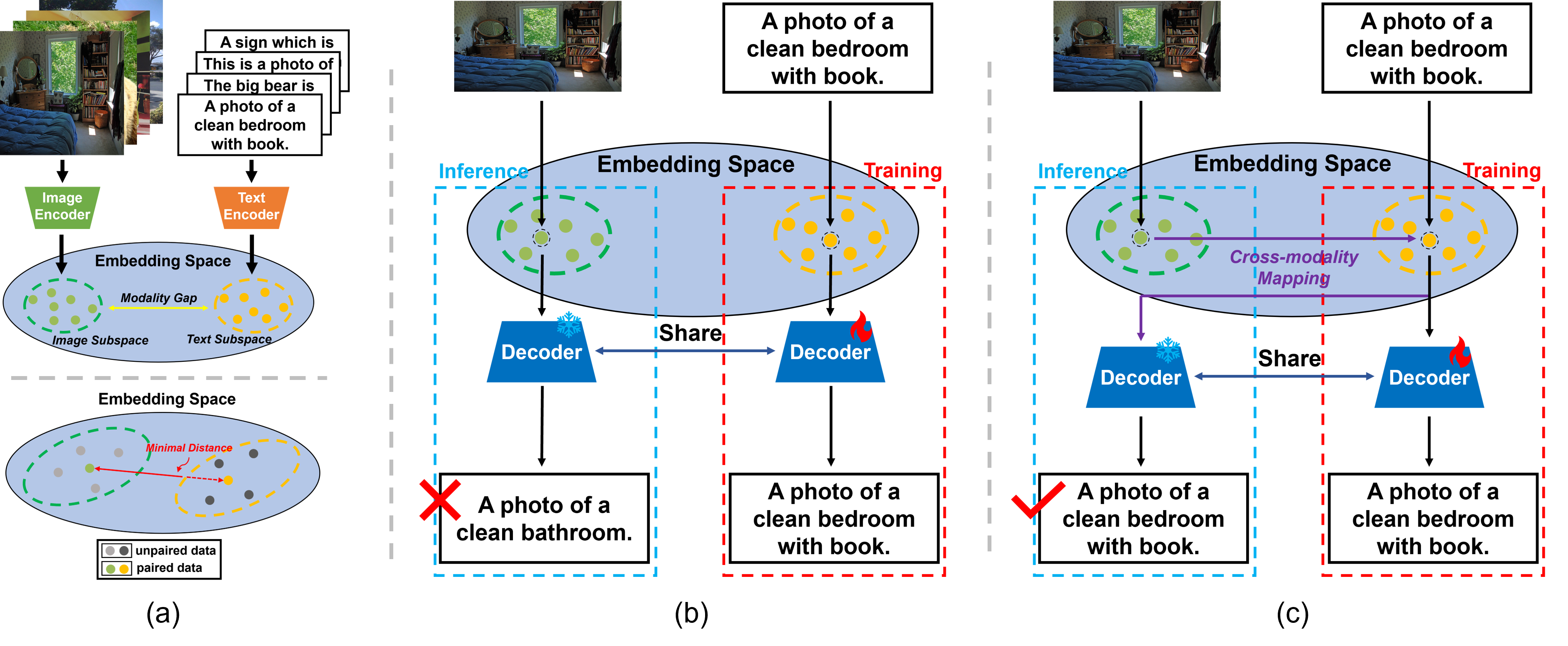}
\caption{\label{fig:figure1} (a) The embedding specificity and cross-modal interaction mode of CLIP. (b) The previous text-only captioning methods based on joint space: in the training phase and the inference phase, the decoder acts in the text and image subspaces, respectively. The modality gap makes the decoder much less effective. (c) Our text-only captioning method \emph{Knight}. Benefiting from cross-modal mapping (in purple arrow), \emph{Knight} alleviates the impact of the modality gap, where the decoder only acts in the text subspaces.}
\vspace{-3mm}
\end{figure*}

The great success of CLIP at the association level has sparked the exploration at the generation level. However, since CLIP does not have a decoder architecture and a pre-training task for generation, it is not competent for the generation-based task. Nevertheless, the dominant performance of large-scale language models such as BERT~\cite{devlin2018bert} and GPT~\cite{radford2019language} makes it possible to decode from the embedding space of CLIP. It is based on this arising the idea of zero-shot generation based on joint space: CLIP encodes pairs of image and text close enough in embedding space~\cite{su2022language,wang2022zero,nukrai2022text}. However,~\cite{liang2022mind} has shown that CLIP encodes images and text into two separate subspaces and there is a significant gap between them as shown in Figure \ref{fig:figure1} (a) top. This means that the decoder is only effective in one modality. When transferring the decoder from one modality to another modality, the modality gap leads to the failure to accurately understand representations as shown in Figure \ref{fig:figure1} (b). 

To eliminate the impact of the modality gap, a major problem is how to establish the transformation relationship between the two modalities. A natural idea is to model the relationship through a large amount of supervised data. However, this requires significant supervised data and training resources~\cite{ramesh2022hierarchical}. We argue that this relationship can be established by an unsupervised method through the association capability of CLIP. Based on this, we propose the \textbf{K}-\textbf{n}earest-ne\textbf{igh}bor Cross-modali\textbf{t}y Mapping (Knight), a text-only captioning method as shown in Figure \ref{fig:figure2}. First, we collect the captions from the image-text and video-text datasets as the corpus. In the training phase, we first select captions from the corpus for training and use the CLIP similarity to retrieve the $k$-nearest-neighbor captions that are most similar to the training captions. Then, we use the CLIP features of the training captions to train the decoder by an autoregression loss. In the inference phase, Knight can be applied to both image and video captioning. For image captioning, we retrieve the $k$-nearest-neighbor captions that are most similar to the inference image. For video captioning, we average the retrieved results for each keyframe to achieve multi-frame input for the video. Knight makes the decoder only act in the text subspace, thus eliminating the effect of the modality gap as shown in Figure \ref{fig:figure1} (c).

We summarize the contributions as follows:
\begin{itemize}
\item We propose a text-only captioning method called Knight based on the unsupervised cross-modal mapping. The method achieves the representation mapping from vision modality to language modality, thus greatly alleviating the impact of the modality gap on cross-modal generation.
\item We compare Knight with the other current zero-shot image and video captioning baselines. Experimental results show that Knight achieves state-of-the-art performance. We explore the possibility of employing CLIP association capacity to address the zero-shot generation-based tasks. 
\end{itemize}

\begin{figure*}[t]
\centering
\includegraphics[width=0.9 \textwidth]{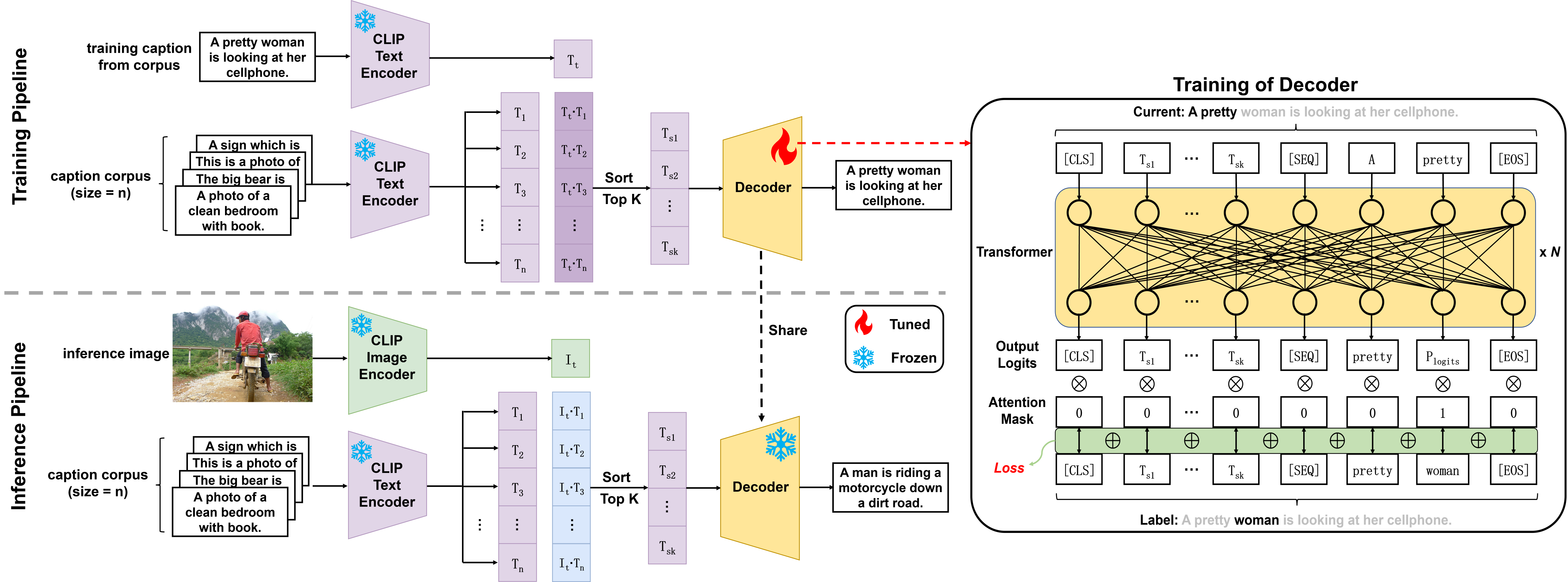}
\caption{\label{fig:figure2} The overview of \emph{Knight}. In the training phase, we first calculate the CLIP similarity to obtain $k$-nearest-neighbor captions from the corpus that are most similar to the training caption. Then, we feed the representations to the decoder for autoregressive training. The training process requires no image or video participation, but only an unsupervised corpus. In the inference phase, we replace the training caption with the inference image and repeat the above steps.}
\vspace{-3mm}
\end{figure*}

\section{Background and Related Work}

\subsection{CLIP}

By contrastive learning on a dataset of 400M (image, text) pairs, CLIP~\cite{radford2021learning} models the association between the images and text. CLIP is widely used for zero-shot tasks such as classification, retrieval, etc~\cite{radford2021learning}. The zero-shot performance is claimed to be close to or even better than fine-tuned models~\cite{radford2021learning}. Many works have applied the zero-shot capacity of CLIP to specific application scenarios such as image segmentation~\cite{xu2021simple}, image generation~\cite{ramesh2022hierarchical}, and object detection~\cite{zhong2022regionCLIP}. 

\subsection{text-only Captioning}

Mainstream methods for captioning tasks are mainly divided into two types: (1) extracting visual features typically using a pre-trained network and training a decoder that produces the final captions~\cite{chen2014learning,chen2017sca,yang2019learning,anderson2018bottom,luo2021dual}; (2) bridging the gap between vision and language by employing pre-training to create a shared latent space of images and text~\cite{lu2019vilbert,laina2019towards,tan2019lxmert,li2020oscar,zhou2020unified,zhang2021vinvl,hu2022scaling}. With the rise of CLIP, recent captioning methods use CLIP for reducing training time~\cite{mokady2021clipcap} and improved captions~\cite{shen2021much,cornia2021universal,kuo2022beyond}.

However, all of the previous works require extensive training and large paired data that are hard to collect. To address this,~\cite{gan2017stylenet} and~\cite{zhao2020memcap} have suggested style-guided captioning, but also employ training over paired data. OpenAI has conducted remarkable work in learning cross-modal patterns with the CLIP trained by a large number of resources~\cite{radford2021learning}. This means that the effective exploitation of the pre-trained knowledge of CLIP can be free from the constraints of supervised data.~\cite{tewel2022zerocap} attempted to generate a caption with the highest CLIP similarity to a given image using a pre-trained language model. However, the pre-trained language model and CLIP's text encoder use different pre-training data and paradigms, making it difficult for the language model to generate high-quality captions that are relevant to images. Researchers realized that additional language data are needed for aligning the embedding space between the language model and CLIP, hence the rise of text-only captioning methods.~\cite{su2022language} made the language model fit the domain of CLIP by fine-tuning it with unsupervised corpus. However, the above works require the language model to provide candidate words and thus prove to have a severe language previous~\cite{wang2022zero}.~\cite{wang2022zero,nukrai2022text} proposed the idea of joint space that argues that CLIP encodes pairs of image and text close enough in embedding space, thus overcoming the problem of language previous by training the decoder in the language modality and transferring it to vision modality in inference phase. 

\begin{table*}[t]
	\centering  
	\renewcommand{\arraystretch}{1.2}
	\setlength{\tabcolsep}{6pt}
	\scalebox{0.9}{
	\begin{tabular}{c c c c c c c c c c c c c c}
		\hline
		\multirow{2}{*}{\textbf{Method}}&\multicolumn{6}{c}{Flickr30k}&\multicolumn{6}{c}{MS-COCO}&\multirow{2}{*}{\textbf{\makecell{Training \\ Params}}}\\
		\cmidrule(lr){2-13}
		&B@1&B@4&M&R-L&CIDEr&SPICE&B@1&B@4&M&R-L&CIDEr&SPICE&\\
		\hline
		\multicolumn{14}{c}{\textit{training-free}}\\
		\hline
		CLIPRe &38.5&5.2&11.6&27.6&10.0&5.7&39.5&4.9&11.4&29.0&13.6&5.3&-\\
		ZeroCap &44.7&5.4&11.8&27.3&16.8&6.2&49.8&7.0&15.4&31.8&34.5&9.2&-\\
		SMs &-&-&-&-&-&-&-&6.9&15.0&34.1&44.5&10.1&-\\
		\hline
        \multicolumn{14}{c}{\textit{text-only training}}\\
        \hline
        MAGIC &44.5&6.4&13.1&31.6&20.4&7.1&56.8&12.9&17.4&39.9&49.3&11.3&345M\\
		CLMs &58.3&16.8&16.2&39.6&22.5&9.8&59.3&15.0&18.7&41.8&55.7&10.9&345M\\
        CapDec &55.5&17.7&20.0&43.9&39.1&9.9&69.2&26.4&25.1&51.8&91.8&11.9&919M\\
		\hline
        Knight (Ours) &\textbf{64.0}&\textbf{22.6}&\textbf{24.0}&\textbf{48.0}&\textbf{56.3}&\textbf{16.3}&\textbf{71.7}&\textbf{27.8}&\textbf{26.4}&\textbf{52.3}&\textbf{98.9}&\textbf{19.6}&771M\\
        \hline
	\end{tabular}
	}
    \caption{Image captioning results of different methods on Flickr30k and MS-COCO, where the B@1, B@4, M, and R-L represent BLEU@1, BLEU@4, METEOR, and Rouge-L respectively.}
    \vspace{-1.5mm}
	\label{tb:image_caption}
\end{table*}

\begin{table*}[h]
    \small
	\centering  
	\renewcommand{\arraystretch}{1.2}
	\setlength{\tabcolsep}{8pt}
	\scalebox{0.9}{
	\begin{tabular}{c c c c c c c c c c c c c}
	\hline
		\multirow{2}{*}{\textbf{Method}}&\multicolumn{6}{c}{MS-COCO $\Longrightarrow$ Flickr30k}&\multicolumn{6}{c}{Flickr30k $\Longrightarrow$ MS-COCO}\\
		\cmidrule(lr){2-7}
		\cmidrule(lr){8-13}
		&B@1&B@4&M&R-L&CIDEr&SPICE&B@1&B@4&M&R-L&CIDEr&SPICE\\
		\hline
		CLIPRe&38.7&4.4&9.6&27.2&5.9&4.2&31.1&3.0&9.9&22.8&8.5&3.9\\        
        MAGIC &46.4&6.2&12.2&31.3&17.5&5.9&41.4&5.2&12.5&30.7&18.3&5.7\\
        CLMs&49.2&10.1&12.5&33.8&12.7&5.7&47.6&7.7&14.9&35.9&38.5&8.2\\        
        CapDec &60.2&17.3&18.6&42.7&35.7&7.2&43.3&9.2&16.3&36.7&27.3&10.4\\   
		\hline
        Knight (Ours)&\textbf{66.0}&\textbf{21.1}&\textbf{22.0}&\textbf{46.3}&\textbf{48.9}&\textbf{14.2}&\textbf{62.1}&\textbf{19.0}&\textbf{22.8}&\textbf{45.8}&\textbf{64.4}&\textbf{15.1}\\
		\hline
	\end{tabular}}
    \caption{Cross-Domain Evaluation. X $\Longrightarrow$ Y  means source domain $\Longrightarrow$ target domain.}
    \vspace{-1.5mm}
	\label{tb:cross_domain_result}
\end{table*}


\subsection{Addressing the Modality Gap}

The previous text-only methods argued that embedded text is relatively close to its corresponding visual embedding~\cite{nukrai2022text}. However,~\cite{liang2022mind} demonstrates that images and text are embedded into two subspaces separately and a significant modality gap is between them as shown in Figure \ref{fig:figure1} (a) top. This gap limits the generation quality of previous methods. Although it is possible to learn the pattern between two modalities with a large amount of data,~\cite{ramesh2022hierarchical} demonstrates that the overhead of this process is huge. Nevertheless, we note that the effective cross-modal interaction of CLIP is under the association of paired data as shown in Figure \ref{fig:figure1} (a) bottom. This makes us realize that although the modality gap makes it hard to achieve cross-modal generation directly, it can be achieved by association indirectly. 

\section{Method}

\subsection{Preliminaries}

\textbf{Notations.} 
We first explain the definition of text-only captioning method and the requirements for the training data. The supervised dataset $\mathcal{D}_{s}$ = \{$(x_1, y_1)$, $\dots$, $(x_n, y_n)$\} consisting of $n$ pairs with images or videos $x_i$ and reference captions $y_i$ = \{$c^1_i$, $\dots$, $c^{|y_i|}_i$\}, where $y_i$ is a set of the captions that describe the $x_i$ from different perspectives, and $c^j_i$ denotes the $j_{th}$ caption of $y_i$. The unsupervised data include unlabeled image or video dataset $\mathcal{D}^{I}_{u}$ = \{$x_1$, $\dots$, $x_i$\} and text datasets $\mathcal{D}^{T}_{u}$ = \{$y_1$, $\dots$, $y_j$\}. Traditional captioning methods use supervised dataset $\mathcal{D}_{s}$ for training, while text-only captioning methods only assume the availability of unlabeled dataset $\mathcal{D}^T_u$.

\textbf{CLIP} is a VLPM with dual-encoder architecture. It consists of two independent encoders for vision and language modalities. Similarities between vision and language representations on large-scale image-text pairs are used to pre-train CLIP, bridging the gap between vision-language semantics in the representation space of CLIP. The similarity is calculated as
\begin{equation}
\begin{aligned}
&I = f_{\textup{V}}(x)\\
&T = f_{\textup{L}}(y)\\
\textup{Sim}(I, T) = &\cos<I, T> = \frac{I}{|I|}\cdot\frac{T}{|T|}  
\end{aligned}
\end{equation}where $f_V$ and $f_L$ are the image encoder and text encoder of CLIP respectively.

\textbf{Unsupervised Language Modelling} is a method for self-supervised training on the unsupervised corpus. It learns the relationship of sentence context by the next-token prediction. The pre-training loss is the Maximum Likelihood Estimation (MLE) that calculated as
\begin{equation}
\mathcal{L}_{\textup{MLE}} = -\frac{1}{|T|}\sum^{|T|}_{i=1} \log M_\theta(T_i|T_1T_2\ldots T_{i-1})
\end{equation}where $\theta$ denotes the parameter that needs to be optimized for the model $M$.

\begin{figure*}[t]
\centering
\includegraphics[width=0.9 \textwidth]{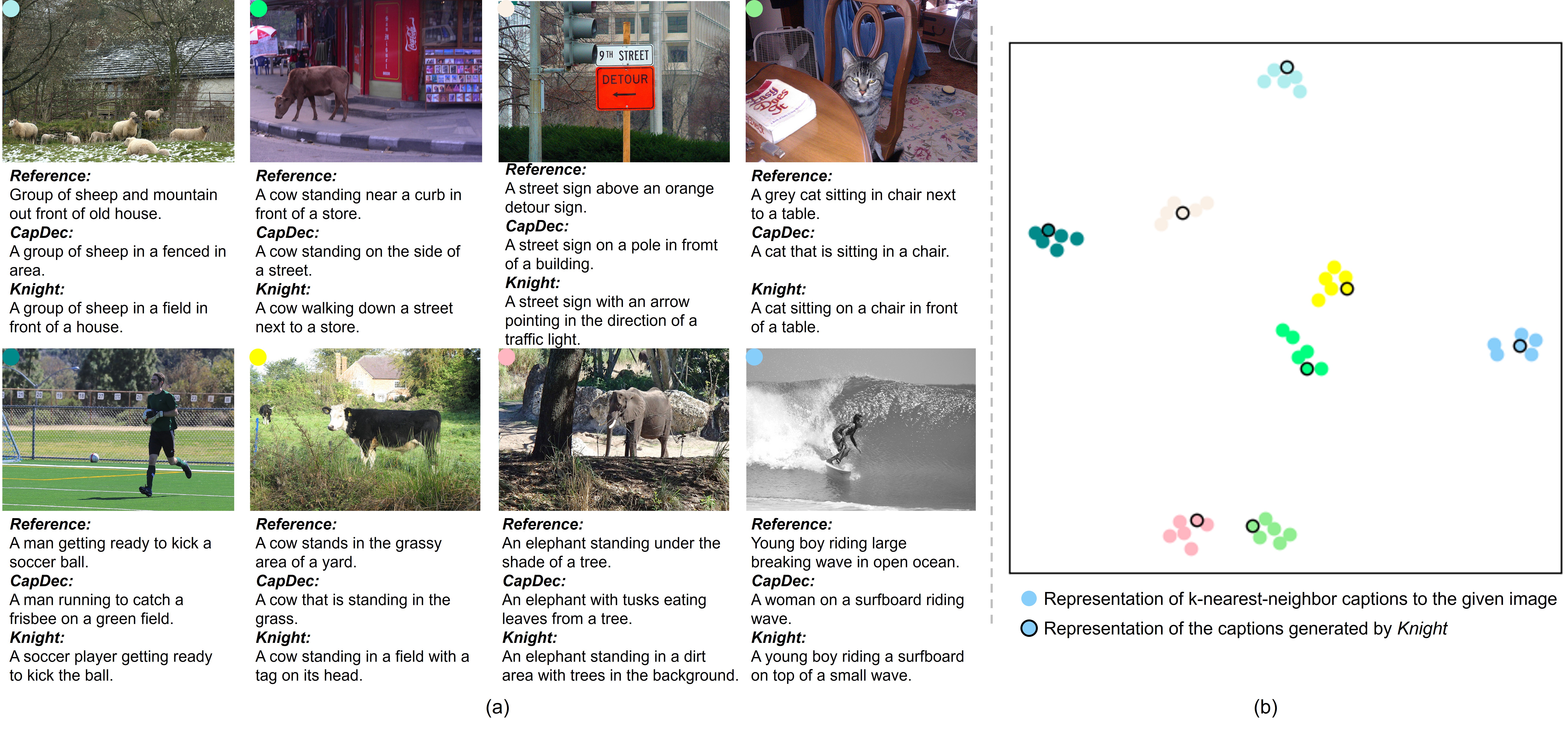}
\caption{\label{fig:figure3} (a): Examples of image captioning generated by \emph{Knight} compared with reference captions. (b): The t-SNE results for the examples in (a). We set $k$ to 5. We take the caption generated by the model back to CLIP and encode it to get the embedding representation of that caption. The solid-colored dots represent the embedding representations of the 5 nearest captions to the given image, and the black-circle dots represent the embedding representations of the captions generated by the model.}
\end{figure*}

\subsection{Knight \label{kinght}}

We consider a training caption $y^{'}$ from $\mathcal{D}^T_u$. The $T_t$ and $T_{1 \sim n}$ calculated as
\begin{equation}
\begin{aligned}
T_t &= f_{\textup{L}}(y^{'}) \\
T_{1 \sim n} = \{f_{\textup{L}}(y_1), &\dots, f_{\textup{L}}(y_n)\} = \{T_1, \dots, T_n\}
\end{aligned}
\end{equation}where $n$ is the size of training corpus. Then, we get the CLIP similarities $S$ between $T_t$ and $T_{1 \sim n}$ calculated as
\begin{equation}
S = \{ \textup{Sim}(T_t, T_1),  \dots, \textup{Sim}(T_t, T_n) \}
\end{equation}
Then we sort the $S$ from large to small and get the $T^{'}$ = \{ $T_{s1}$, \dots, $T_{sk}$ \}, where $T_{si}$ is the $i$-th largest of the sorted result and $k$ is a hyperparameter. Finally, we use the $T^{'}$ for autoregression training as following equation
\begin{equation} 
\label{e5}
\mathcal{L}_{\textup{MLE}} = -\frac{1}{|T^{'}|}\sum^{|T^{'}|}_{i=1} \log M_\theta(T_i|T_{s1} \ldots T_{sk}, T_1 \ldots T_{i-1})
\end{equation}

\subsubsection{Image Captioning}

After training, we fix the parameters of the decoder. In the inference phase, we replace the input of the training caption with the inference image $x$ and get the $I_t$ = $f_{\textup{V}}(x)$. We get the $S$ as the following equation
\begin{equation}
S = \{ \textup{Sim}(\bm{I_t}, T_1),  \dots, \textup{Sim}(\bm{I_t}, T_n) \}
\end{equation}
Finally, we follow the training process to obtain $T^{'}$ and feed it into the decoder to get the generated captions as the equation (\ref{e5}).

\subsubsection{Video Captioning}

The above process of image captioning can be understood as a process of filtering and regrouping the information from $k$-nearest-neighbor captions. This process is similar to the generation of video caption: the decoder establishes the connection between the preceding and following frames and reasons out the combined information of these frames. Therefore, we argue that Knight can be transferred to video captioning.

Compared with the image, video is a set of multiple frames. We first extract the keyframes $x$ = \{ $x_1$, \ldots, $x_m$ \} of the inference video $x$, where $m$ is the number of keyframes. Then, we get the $I_t^i$ = $f_{\textup{V}}(x_i)$, where $x_i$ is the $i$-th keyframe.
For each keyframe, we get the $S_i$ as the following equation
\begin{equation}
S_i = \{ \textup{Sim}(\bm{I_t^i}, T_1), \dots, \textup{Sim}(\bm{I_t^i}, T_n) \}
\end{equation}
We sort each $S_i$ and select the $k$ representations with the greatest similarity thus obtaining $T_i^{'}$. Finally, we get the $T^{'}$ for video as the following equation
\begin{equation}
T^{'} = \{ T_{s1}, \dots, T_{sm} \} = \{ mean(T_1^{'}), \dots, mean(T_m^{'}) \}
\end{equation}
Finally, we feed $T^{'}$ into the decoder to generate the captions as the equation (\ref{e5}).

\section{Experiment}

\subsection{Experimental Settings}

\textbf{Evaluation Benchmarks.}
For the image captioning task, we conduct experiments on two widely used benchmarks: Flickr30k~\cite{plummer2015flickr30k} and MS-COCO~\cite{lin2014microsoft,chen2015microsoft}. And we set up the training, validation, and test splits according to the protocols provided by Karpathy et al for both datasets~\cite{karpathy2015deep}. For the video captioning task, we choose two video datasets: MSR-VTT~\cite{xu2016msr} and MSVD~\cite{wu2017deep}. We use the captions in the training set as the training corpus and evaluate the methods on the test set, as in other text-only captioning works.

\noindent \textbf{Implementation Details.}
For CLIP, we choose the Resnet50x64 architecture which encodes each image as a 1024-dimension vector. For the decoder, we choose the large vision of GPT-2~\cite{radford2019language} with a 1280-dimension embedding space. To align CLIP and decoder on the representation layer, we use a 3-layer MLP that transforms the representation of CLIP into 1280 dimensions. We optimize the decoder with the Adam optimizer~\cite{kingma2014adam} and a learning rate of 1e-6. Since no vision modality is involved, the training process is computationally negligible, i.e., less than 6 hours with 1 Tesla A100 GPU. In the inference phase, we use beam search, where the branches of the beam are chosen as 5 which is as same as other methods. For the acquiring of video keyframes, we choose the isometric sampling.

\noindent \textbf{Baselines.} For image captioning, we include several zero-shot methods as our baselines. First, we compare with a training-free method, called CLIPRe~\cite{su2022language}. Given an image, it retrieves the most related caption from the training corpus based on the image-text similarity as measured by CLIP. Then, we compare two training-free methods with the language model ZeroCap~\cite{tewel2022zerocap} and SMs~\cite{zeng2022socratic}. Finally, we compare with text-only methods MAGIC~\cite{su2022language}, CLMs~\cite{wang2022zero}, and current state-of-the-art method CapDec~\cite{nukrai2022text}. For video captioning, we compare it with the state-of-the-art method EPT~\cite{tewel2022zero}. EPT is a zero-shot video captioning method based on evolving pseudo-tokens. And we also adapt the baselines of image captioning to video captioning by referring to~\cite{tewel2022zero}. These methods use the average features of video keyframes as input.

\noindent \textbf{Evaluation Metrics.}
Following the common practice in the literature, we perform an evaluation using BLEU-1 (B@1), BLEU-4 (B@4)~\cite{papineni2002bleu}, METEOR (M)~\cite{denkowski2014meteor}, ROUGE-L (R-L)~\cite{lin2004automatic}, CIDEr~\cite{vedantam2015cider}, and SPICE~\cite{anderson2016spice}.

\begin{table*}[t]
	\centering  
	\renewcommand{\arraystretch}{1.2}
	\setlength{\tabcolsep}{8pt}
	\scalebox{0.9}{
	\begin{tabular}{c c c c c c c c c c c c c}
		\hline
		\multirow{2}{*}{\textbf{Method}}&\multicolumn{6}{c}{MSR-VTT}&\multicolumn{6}{c}{MSVD}\\
		\cmidrule(lr){2-13}
		&B@1&B@4&M&R-L&CIDEr&SPICE&B@1&B@4&M&R-L&CIDEr&SPICE\\
		\hline
		\multicolumn{13}{c}{\textit{full-supervised training}}\\
		\hline
		VNS-GRU&-&45.3&29.9&64.7&53.0&-&-&66.5&42.1&79.7&121.5&-\\
		SemSynAN &-&46.4&30.4&46.7&51.9&-&-&64.4&41.9&79.5&111.5&-\\
		\hline
        \multicolumn{13}{c}{\textit{unsupervised training}}\\
        \hline
        ZeroCap*$^\dag$ &-&2.3&12.9&30.4&5.8&-&-&2.9&16.3&35.4&9.6&-\\
        MAGIC*  &22.3&5.5&13.3&35.4&7.4&4.2&24.7&6.6&16.1&40.1&14.0&2.9\\
		CLMs* &25.7&6.2&17.8&15.7&10.1&6.5&26.9&7.0&16.4&44.3&20.0&3.1\\
        CapDec* &30.2&8.9&23.7&17.2&11.5&5.9&33.1&7.9&23.3&25.2&34.5&3.2\\
        ETP$^\dag$  &-&3.0&14.6&22.7&11.3&-&-&3.0&17.8&31.4&17.4&-\\
		\hline
        Knight (Ours) &\textbf{72.6}&\textbf{25.4}&\textbf{28.0}&\textbf{50.7}&\textbf{31.9}&\textbf{8.5}&\textbf{73.1}&\textbf{37.7}&\textbf{36.1}&\textbf{66.0}&\textbf{63.8}&\textbf{5.0}\\
        \hline
        \multicolumn{13}{l}{*The method is adapted from zero-shot image captioning to zero-shot video captioning.}\\
        \multicolumn{13}{l}{$^\dag$The method is training-free.}\\
	\end{tabular}}
    \caption{Video captioning results of different methods on MSR-VTT and MSVD.}
    \vspace{-1.5mm}
	\label{tb:video_caption}
\end{table*}

\begin{figure*}[t]
\centering
\includegraphics[width=0.9 \textwidth]{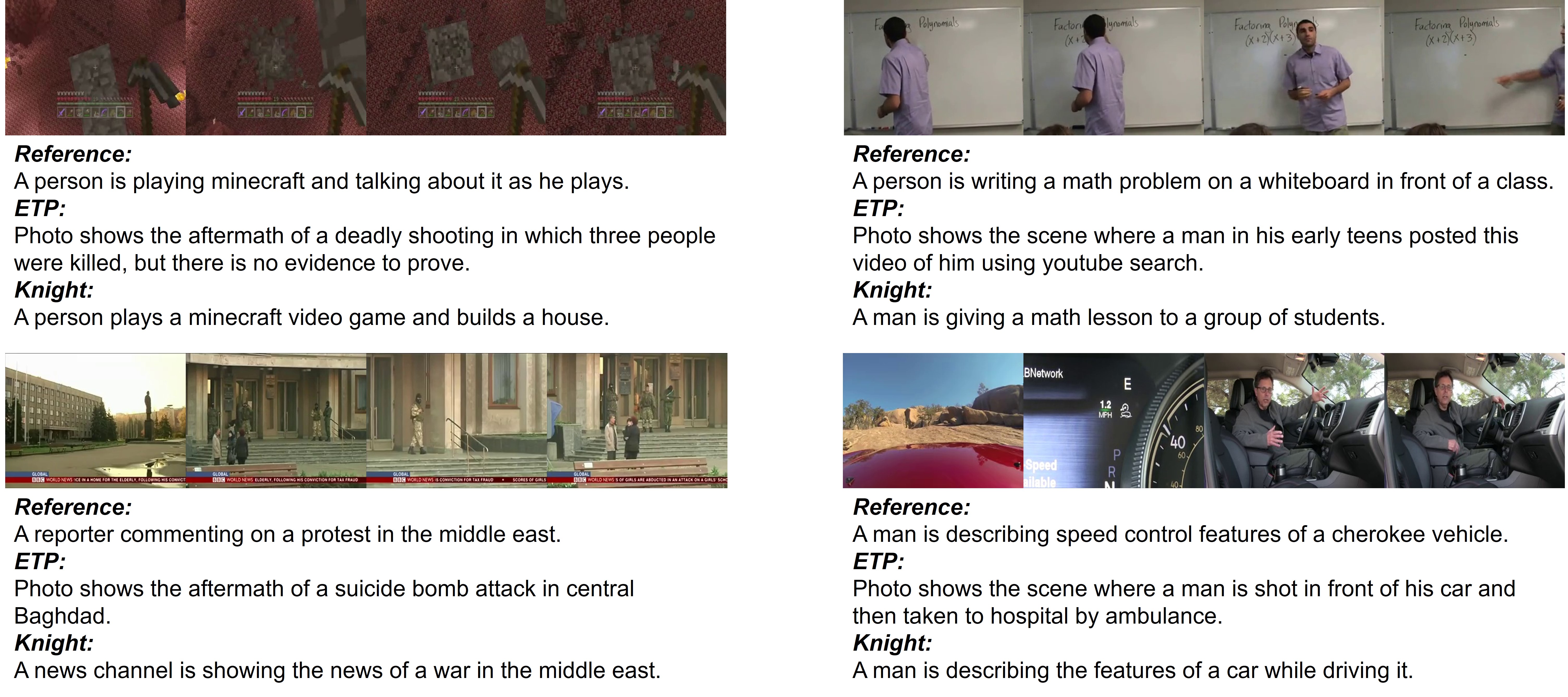}
\caption{\label{fig:figure4} Examples of video captioning generated by \emph{Knight} with reference captions.}
\end{figure*}

\subsection{Performance Comparison}

\subsubsection{Image Captioning}

The results of image captioning are shown in Table \ref{tb:image_caption}. We see that Knight achieves the best performance in all 12 evaluation metrics. It is worth noting that compared to the current state-of-the-art method CapDec, Knight achieves significant performance improvement using fewer training parameters. This is because CapDec is based on joint space and alleviates the impact of the modality gap by noise interference. However, it is not reasonable to model the modality gap by random Gaussian noise because the offset direction of the modality gap is directional, while random Gaussian noise is non-directional. Compared with CapDec, Knight utilizes the similarity mapping that is best suited for CLIP. The direction of this mapping is consistent with the offset direction of the modality gap.

To test the generalization ability of Knight, we conduct a cross-domain experiment. Specifically, we apply the text-only training on the corpus of the source domain (e.g., MS-COCO) to perform inference on the test set of the target domain (e.g., Flickr30k). The results of the cross-domain experiment are shown in Table \ref{tb:cross_domain_result}. From the results, we can see that Knight has a significant advantage in generalization ability.

We show some examples generated by Knight compared with CapDec in Figure \ref{fig:figure3} (a). We see that the caption generated by Knight corresponds well to the content in the image, both in the foreground and in the background. We understand the reason for the high-quality generation of Knight by visualizing the results in Figure \ref{fig:figure3} (b). We reduce the embedding space representations of the $k$-nearest-neighbor captions to a 2-dimensional representation by t-SNE (solid-colored dots), while bringing the generated caption back to the embedding space to obtain the corresponding representation (black-circle dots). We can see that the $k$-nearest-neighbor captions of different images are significantly distinguished on the embedding space, which indicates that Knight has a good discriminatory ability for the different concepts.

\subsubsection{Video Captioning}

The results of video captioning are shown in Table \ref{tb:video_caption}. From the results of the methods adapted from image captioning, we see that these methods do not apply to video captioning. This is because the input to these methods is a representation of only a single image or text. When multiple keyframes from a video are used as input, these methods need to fuse the representations of multiple frames into a single representation. This not only causes the loss of features, but also fails to model the relationship between different keyframes, and thus fails to inference about the behavior of the whole video. 

From the results of ETP, we see that although it is a method designed for video captioning, it still does not work well. This is for the same reason that ZeroCap in image captioning fails to generate high-quality captions: the CLIP used for matching computation cannot be aligned with the language model used for a generation without text training. Finally, although there are still gaps in Knight compared to the full-supervised method, Knight achieves the best in all 12 evaluation metrics in the unsupervised method and is far ahead of the other methods.

In the example of Figure \ref{fig:figure0}, although Knight does not use the image of \emph{Squidward} for training, it successfully maps to \emph{Squidward} by unsupervised cross-modal mapping of CLIP. In Figure \ref{fig:figure4}, we show more examples. It is clear from the results that Knight has a strong capacity to recognize the concept of open worlds. For example, the game name \emph{Minecraft} in the first example, \emph{math} in the second example, and the \emph{Middle East} in the third example.

\subsection{Discusions}

\noindent \textbf{\emph{A. Are the results of Knight due to the captions in the training corpus that are similar to the inference images or videos?}}

We compare CLIPRe (Table \ref{tb:image_caption}) to analyze the issue. Given an image, CLIPRe retrieves the most related caption from the training corpus based on the image-text similarity as measured by CLIP. From the result, we see that a better result cannot be achieved by only retrieving from the corpus, although the Flickr30K and MS-COCO are large enough. This indicates that our method is not a retrieval method but a generated method.

\begin{figure}[t]
\centering
\includegraphics[width=0.475 \textwidth]{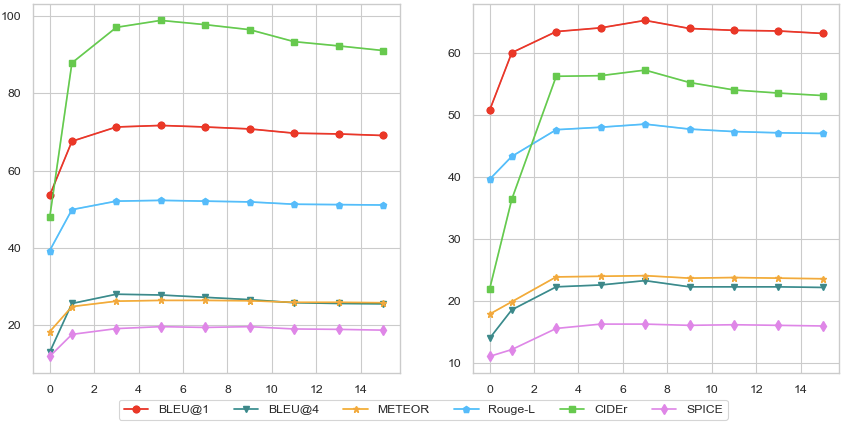}
\caption{\label{fig:figure5} The image captioning results of \emph{Knight} under different $k$ value settings on MS-COCO (left) and Flickr30K (right), where the horizontal coordinate is the value of $k$ and the vertical coordinate is the value of each metric.}
\end{figure}

~\\
\noindent \textbf{\emph{B. Does irrelevant content in the $k$-nearest-neighbor captions of an image impact the generation quality?}}

From the result in Figure \ref{fig:figure3} (b), we can see that although the generated caption is strongly associated with the $k$-nearest-neighbor captions of the image, there is still a distinction between the two. This indicates that the decoder is not overly dependent on a caption and generates according to its feature, but is a reorganization of the information in the $k$-nearest-neighbor captions. Thus even if there is irrelevant information in one of the $k$-nearest-neighbor captions, the decoder does not rely on this information completely.

~\\
\noindent \textbf{\emph{C. What effect does the choice of $k$ have on the performance of Knight?}}

We set different values of $k$ to explore this issue. The results on image captioning are shown in Figure \ref{fig:figure5} and the results on video captioning are shown in appendix. It is worth noting that $k$ equals 0 means that the representation of the inference image is used to generate directly. First, the performance at $k$ of 0 is not good. This confirms the existence of the modality gap. Then, when $k$ is 1, i.e., the nearest-neighbor mapping, the performance is still not optimal. Finally, as $k$ increases, the performance of Knight no longer changes significantly after rising to a certain level. Although the value of $k$ increases, there is a limited number of captions associated with inference images in the corpus. An excessively large $k$ is not only detrimental to good performance but also introduces noise.

\begin{figure}[t]
\centering
\includegraphics[width=0.475 \textwidth]{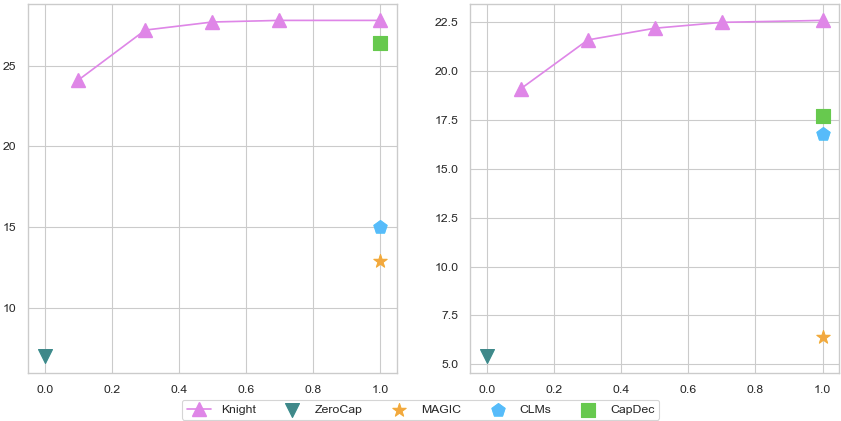}
\caption{\label{fig:figure6} The results of B@4 of \emph{Knight} with different proportions of training corpus compared with other methods on MS-COCO (left) and Flickr (right), where the horizontal coordinate is the proportions of training corpus and the vertical coordinate is the value of B@4.}
\end{figure}

~\\
\noindent \textbf{\emph{D. Does Knight show a significant performance decline when the size of the training corpus is reduced?}}

We set different sizes of the training corpus of Knight to explore this issue. The results of B@4 are shown in Figure \ref{fig:figure6} and the results of other metrics are shown in the appendix. We observe that the performance of Knight does decline when the size of the training corpus is reduced. However, it is worth noting that Knight does not suffer from a catastrophic performance drop even with just 10$\%$ of the training corpus. Compared to other methods, Knight still has significant advantages.

\section{Conclusion}

In this paper, we explore how to achieve generation by utilizing the zero-shot capability of CLIP. To alleviate the negative impact of the modality gap on the text-only captioning method, we propose the association-to-generation method, Knight. Unlike previous works, in the inference phase, we represent a given image as $k$-nearest-neighbor captions by computing the similarity between the inference image or video and the captions in the corpus. This unifies the decoding range in the training and inference phases and makes the generated captions independent of the modality gap. Experimental results show that Knight achieves state-of-the-art performances on both text-only image and video captioning.

\clearpage
\bibliographystyle{named}
\bibliography{ijcai23}
\clearpage

\appendix

\section{Training Data}

\begin{table}[t]
	\centering  
	\renewcommand{\arraystretch}{1.5}
	\setlength{\tabcolsep}{9pt}
	\scalebox{1}{
	\begin{tabular}{c  c  c}
		\hline
		Dataset&Training Captions&Inference Images\\
		\hline
		Flickr30K&144992&1000\\
		MS-COCO&566720&5000\\
		MSR-VTT&140192&2990\\
		MSVD&48774&670\\
		\hline
	\end{tabular}
	}
    \caption{Detailed training data information for different datasets.}
    \vspace{-1.5mm}
	\label{tb:train data}
\end{table}

\begin{table}[t]
	\centering  
	\renewcommand{\arraystretch}{1.2}
	\setlength{\tabcolsep}{9pt}
	\scalebox{1}{
	\begin{tabular}{c c}
		\hline
        Backbone&Resnet50x64\\
		Embedding Size&1024\\
		Architecture&Encoder\\
		\hline
	\end{tabular}
	}
    \caption{The details of CLIP used in this paper.}
    \vspace{-1.5mm}
	\label{tb:CLIP}
\end{table}

\begin{table}[t]
	\centering  
	\renewcommand{\arraystretch}{1.2}
	\setlength{\tabcolsep}{9pt}
	\scalebox{1}{
	\begin{tabular}{c  c}
		\hline
        Backbone&Transformer\\
        Layer Number&36\\
        Head Number&20\\
        Activation Function&GELU\\
		Embedding Size&1280\\
        Tokenizer&GPT-2 Large\\
		Architecture&Decoder\\
        \hline
	\end{tabular}
	}
    \caption{The details of GPT-2 used in this paper.}
    \vspace{-1.5mm}
	\label{tb:GPT}
\end{table}

\begin{figure}[t]
\centering
\includegraphics[width=0.45 \textwidth]{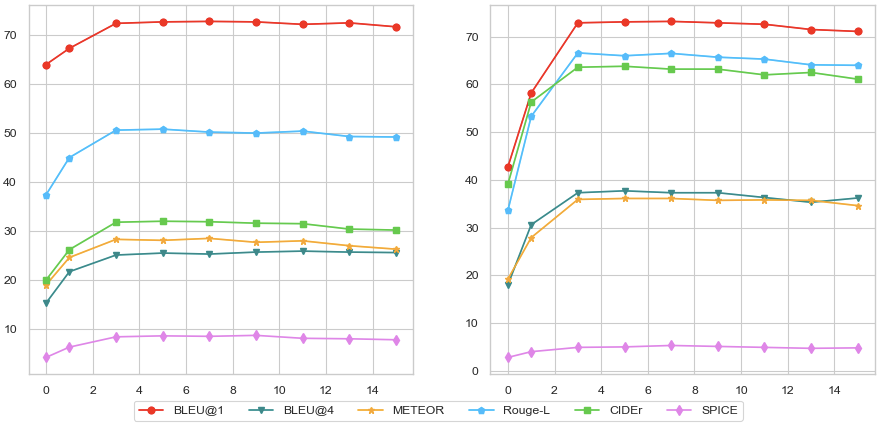}
\caption{\label{fig:figure1} The image captioning results of \emph{Knight} under different $k$ value settings on MSR-VTT (left) and MSVD (right), where the horizontal coordinate is the value of $k$ and the vertical coordinate is the value of each metric.}
\end{figure}

\begin{figure}[t]
\centering
\includegraphics[width=0.45 \textwidth]{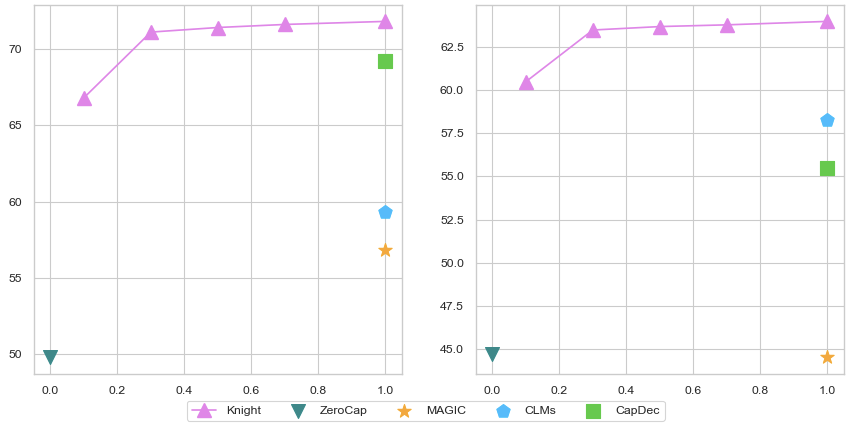}
\caption{\label{fig:figure2} The results of B@1 of \emph{Knight} with different proportions of training corpus compared with other methods on MS-COCO (left) and Flickr (right), where the horizontal coordinate is the proportions of training corpus and the vertical coordinate is the value of B@1.}
\end{figure}

\begin{figure}[t]
\centering
\includegraphics[width=0.45 \textwidth]{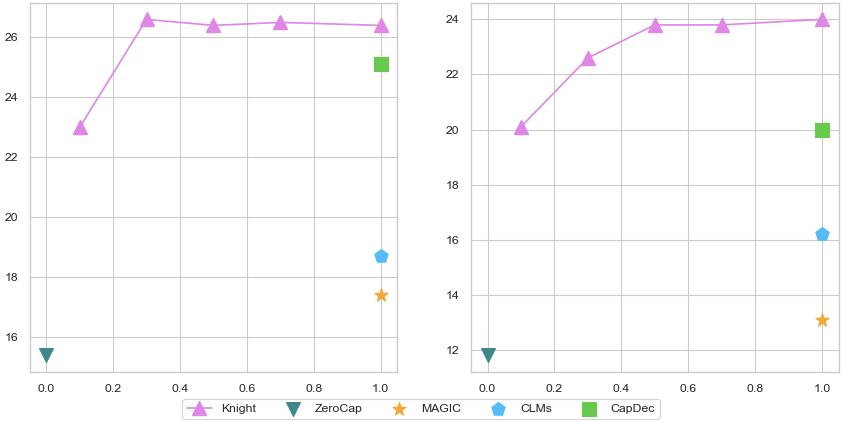}
\caption{\label{fig:figure3} The results of METEOR of \emph{Knight} with different proportions of training corpus compared with other methods on MS-COCO (left) and Flickr (right), where the horizontal coordinate is the proportions of training corpus and the vertical coordinate is the value of METEOR.}
\end{figure}

\begin{figure}[t]
\centering
\includegraphics[width=0.45 \textwidth]{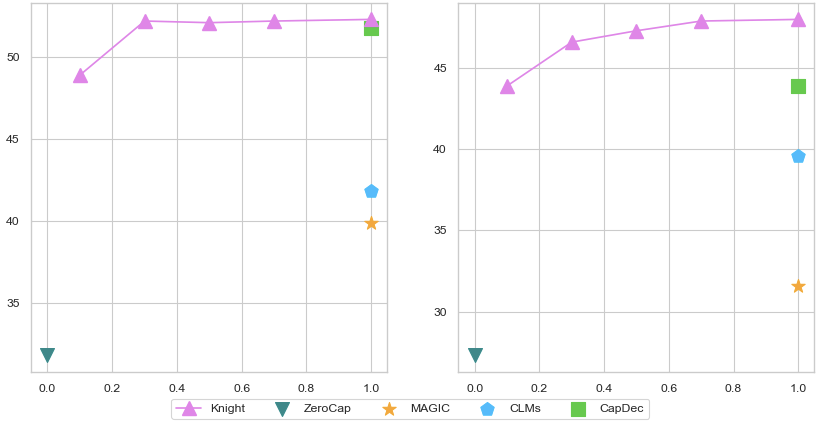}
\caption{\label{fig:figure4} The results of Rouge-L of \emph{Knight} with different proportions of training corpus compared with other methods on MS-COCO (left) and Flickr (right), where the horizontal coordinate is the proportions of training corpus and the vertical coordinate is the value of Rouge-L.}
\end{figure}

\begin{figure}[t]
\centering
\includegraphics[width=0.45 \textwidth]{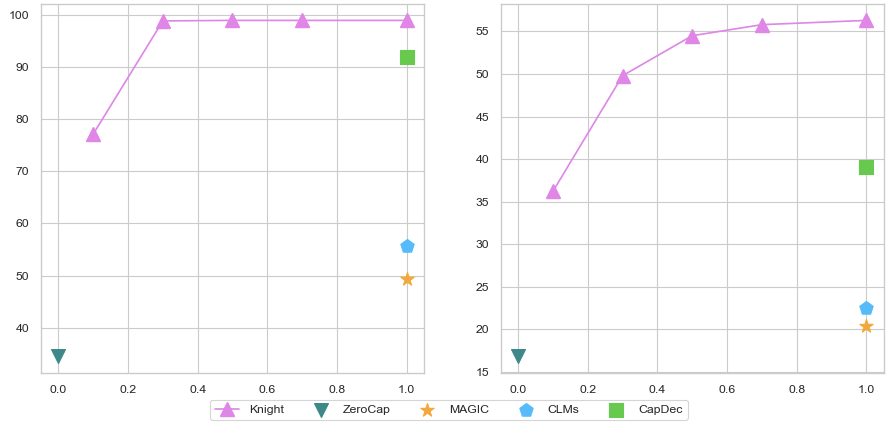}
\caption{\label{fig:figure5} The results of CIDEr of \emph{Knight} with different proportions of training corpus compared with other methods on MS-COCO (left) and Flickr (right), where the horizontal coordinate is the proportions of training corpus and the vertical coordinate is the value of CIDEr.}
\end{figure}

\begin{figure}[t]
\centering
\includegraphics[width=0.45 \textwidth]{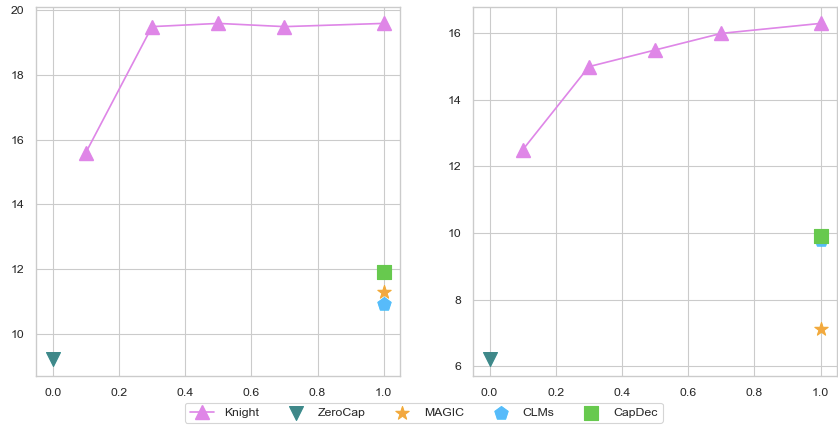}
\caption{\label{fig:figure6} The results of SPICE of \emph{Knight} with different proportions of training corpus compared with other methods on MS-COCO (left) and Flickr (right), where the horizontal coordinate is the proportions of training corpus and the vertical coordinate is the value of SPICE.}
\end{figure}

\begin{figure}[t]
\centering
\includegraphics[width=0.45 \textwidth]{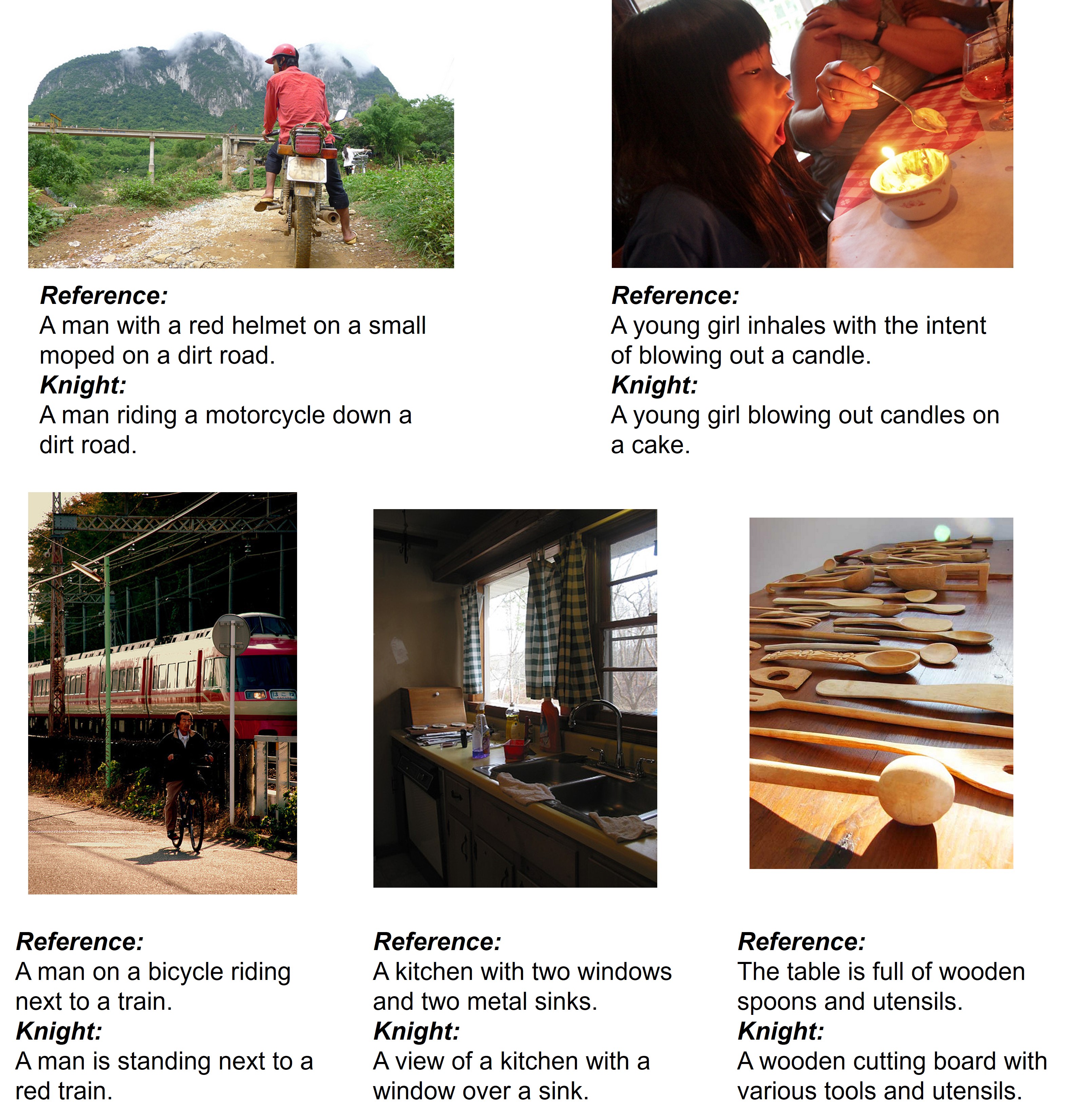}
\caption{\label{fig:figure7} The generated results of Knight on the first 5 inference examples of MS-COCO.}
\end{figure}

\begin{figure}[t]
\centering
\includegraphics[width=0.45 \textwidth]{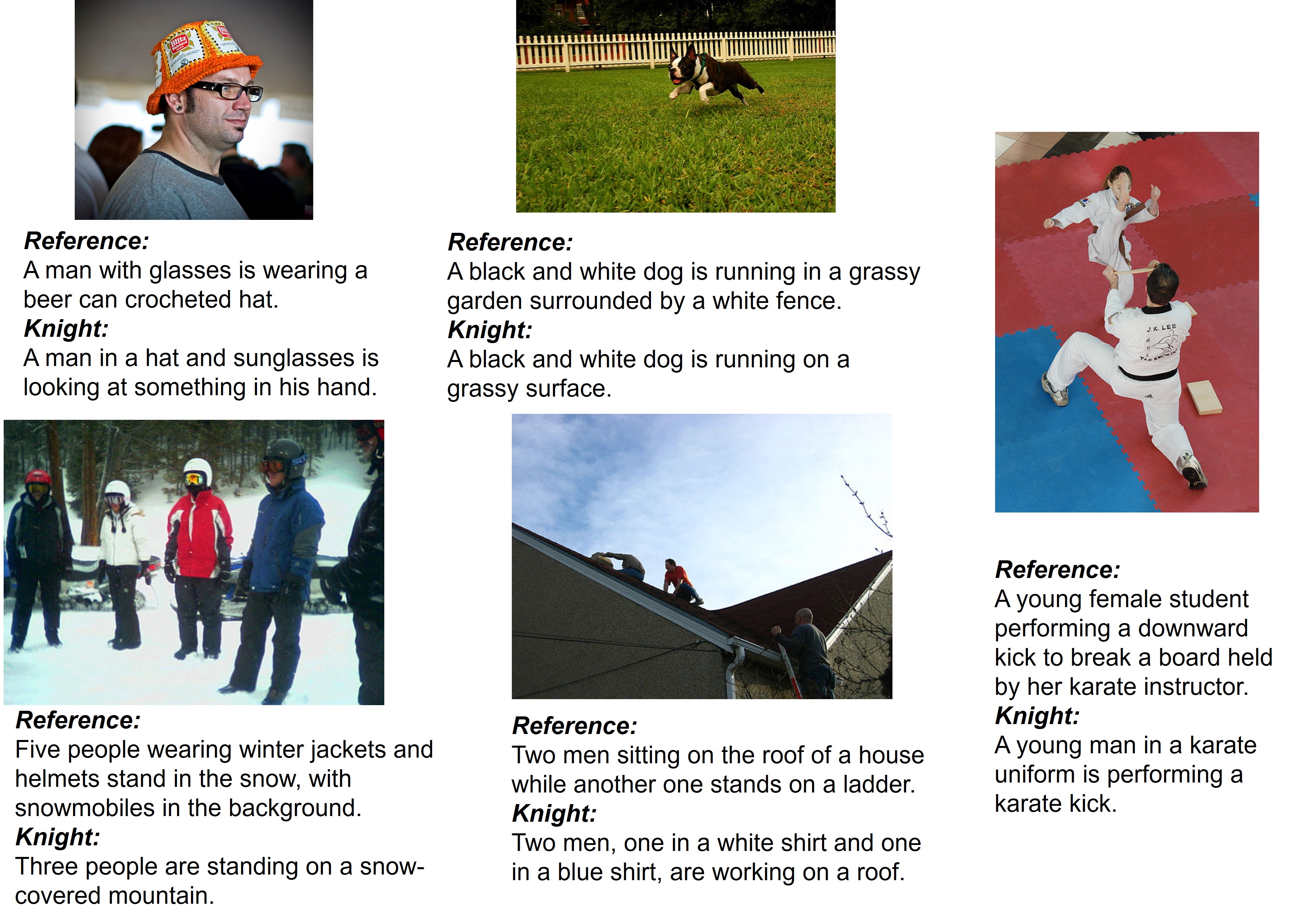}
\caption{\label{fig:figure8} The generated results of Knight on the first 5 inference examples of Flickr30K.}
\end{figure}

We use a total of 4 datasets: Flickr30K, MS-COCO, MSR-VTT, and MSVD. Detailed training data information for different datasets is shown in Table \ref{tb:train data}. We collect the captions from the training splits of each dataset as the training corpus and the inference images/videos from the test splits of each dataset.

\section{Model Details}

We use CLIP as encoder and gpt2-large as decoder.The detailed information is shown in Table \ref{tb:CLIP} and \ref{tb:GPT}, respectively. Since the embedding space sizes of the two are different, we use a 3-layer MLP to implement a 1024-dimensional to 1280-dimensional mapping.

\section{Supplementary Experimental Results}

In the body section we show the effect of training corpus size on the B@4 results. The results of the other metrics (B@1, METEOR, Rouge-L, CIDEr, SPICE) are shown in Figures \ref{fig:figure2}, \ref{fig:figure3}, \ref{fig:figure4}, \ref{fig:figure5}, and \ref{fig:figure6} in this section.

\section{More Example}
We show more examples in both MS-COCO and Flickr datasets. We selected the top 5 inference examples from each dataset for display.

\end{document}